\documentclass[11pt,letterpaper]{article}
\usepackage[letterpaper]{geometry}
\usepackage{acl2012}
\usepackage{times}
\usepackage{latexsym}
\usepackage[reqno]{amsmath}
\usepackage{multirow}
\usepackage{url}

\makeatletter
\newcommand{\@BIBLABEL}{\@emptybiblabel}
\newcommand{\@emptybiblabel}[1]{}
\makeatother
\usepackage[hidelinks]{hyperref}

\usepackage{graphicx}
\usepackage{subfigure}

\usepackage{mathtools}
\usepackage{relsize} 

\usepackage[framemethod=TikZ]{mdframed}

\usepackage{color,soul}

\graphicspath{ {./figures} }

\usepackage{relsize} 

\usepackage{fixltx2e}

\usepackage[compact]{titlesec}
\titlespacing{\section}{0pt}{2ex}{1ex}
\titlespacing{\subsection}{0pt}{1ex}{0ex}
\titlespacing{\subsubsection}{0.2pt}{0.5ex}{0ex}

\usepackage[font=small,skip=0pt]{caption} 

\usepackage{array}
\newcolumntype{P}[1]{>{\centering\arraybackslash}p{#1}} 
\newcolumntype{M}[1]{>{\centering\arraybackslash}m{#1}} 

\usepackage{enumitem}

\title{Modeling Semantic Expectation: \\ Using Script Knowledge for Referent Prediction }

\author{   Ashutosh Modi$^{1,3}$   Ivan Titov$^{2,4}$  Vera Demberg$^{1,3}$   Asad Sayeed$^{1,3}$   Manfred Pinkal$^{1,3}$ 
\\
$^{1}$  {\tt \{ashutosh,vera,asayeed,pinkal\}@coli.uni-saarland.de} \\ 
$^{2}$  {\tt titov@uva.nl} \\
$^{3}$ Universit{\"a}t des Saarlandes, Germany \\
$^{4}$  ILLC, University of Amsterdam, the Netherlands  \\
}

\date{}

\begin{document}
\maketitle
\begin{abstract}
Recent research in psycholinguistics has provided increasing evidence that humans predict upcoming content. Prediction also affects perception and might be a key to robustness in human language processing. In this paper, we investigate the factors that affect human prediction by building a computational model that can predict upcoming discourse referents based on linguistic knowledge alone vs.~linguistic knowledge jointly with common-sense knowledge in the form of scripts. We find that script knowledge significantly improves model estimates of human predictions. In a second study, we test the highly controversial hypothesis that predictability influences referring expression type but do not find evidence for such an effect.

\end{abstract}


\section{Introduction}
Being able to anticipate upcoming content is a core property of human language processing \cite{kutas2011look,kuperberg2016we} that has received a lot of attention in the psycholinguistic literature in recent years. Expectations about upcoming words help humans comprehend language in noisy settings and deal with ungrammatical input. In this paper, we use a computational model to address the question of how different layers of knowledge (linguistic knowledge as well as common-sense knowledge) influence human anticipation. 

Here we focus our attention on semantic predictions of \textit{discourse referents} for upcoming noun phrases. This task is particularly interesting because it allows us to separate the semantic task of anticipating an intended referent and the processing of the actual surface form. For example, in the context of \textit{I ordered a medium sirloin steak with fries. Later, the waiter brought \ldots}, 
there is a strong expectation of a specific discourse referent, i.e., the referent introduced by the object NP of the preceding sentence, 
while the possible referring expression could be either \textit{the steak I had ordered}, \textit{the steak}, \textit{our food}, or \textit{it}.
Existing models of human prediction are usually formulated using the information-theoretic concept of \textit{surprisal}. In recent work, however, surprisal is usually not computed for DRs, which represent the relevant semantic unit, but for the surface form of the referring expressions, even though there is an increasing amount of literature suggesting that human expectations at different levels of representation have separable effects on prediction and, as a consequence, that 
the modelling of only one level (the linguistic surface form) is insufficient \cite{kuperberg2016we,kuperberg2016separate,zarcone2016salience}.
The present model addresses this shortcoming 
by explicitly modelling and representing common-sense knowledge and conceptually separating the semantic (discourse referent) and the surface level (referring expression) expectations. 

Our discourse referent prediction task is related to the NLP task of coreference resolution, but it substantially differs from that task in the following ways: 1) we use only the incrementally available left context, while coreference resolution uses the full text; 2) coreference resolution tries to identify the DR for a given target NP in context, while we look at the expectations of DRs based only on the context before the target NP is seen.

The distinction between referent prediction and prediction of referring expressions also allows us to study a closely related question in natural language generation: the choice of a type of referring expression based on the predictability of the DR that is intended by the speaker. This part of our work is inspired by a referent guessing experiment by Tily and Piantadosi~\shortcite{tily2009refer}, who showed that highly predictable referents were more likely to be realized with a pronoun than unpredictable referents, which were more likely to be realized using a full NP. The effect they observe is consistent with a Gricean point of view, or the principle of uniform information density (see Section \ref{sec:uid}). However, Tily and Piantadosi do not provide a computational model for estimating referent predictability. Also, they do not include selectional preference or common-sense knowledge effects in their analysis. 

We believe that \textit{script knowledge}, i.e., common-sense knowledge about everyday event sequences, represents a good starting point for modelling conversational anticipation. This type of common-sense knowledge includes temporal structure which is particularly relevant for anticipation in continuous language processing.
Furthermore, our approach can build on progress that has been made in recent years in methods for acquiring large-scale script knowledge; see Section \ref{sec:scripts}.
Our hypothesis is that script knowledge may be a significant factor in human anticipation of discourse referents. Explicitly modelling this knowledge will thus allow us to produce more human-like predictions. 

Script knowledge enables our model to generate anticipations about discourse referents that have already been mentioned in the text, as well as anticipations about textually new discourse referents which have been activated due to script knowledge. By modelling event sequences and event participants, our model captures many more long-range dependencies than normal language models are able to. As an example, consider the following two alternative text passages:

\textit{We got seated, and had to wait for 20 minutes. Then, the waiter brought the ...}

 \textit{We ordered, and had to wait for 20 minutes. Then, the waiter brought the ...}
 
 Preferred candidate referents for the object position of \textit{the waiter brought the ...} are instances of the \textit{food}, \textit{menu}, or \textit{bill} participant types. In the context of the alternative preceding sentences, there is a strong expectation of  instances of a \textit{menu} and a \textit{food} participant, respectively. 
\definecolor{orange}{RGB}{255,150,0}
\definecolor{purp}{RGB}{127,0,255}
\definecolor{somecolor}{RGB}{50,200,200} 
\definecolor{somecolor2}{RGB}{255,0,255} 
\definecolor{somecolor3}{RGB}{200,100,100} 
\definecolor{somecolor4}{RGB}{100,100,100} 
\definecolor{somecolor5}{RGB}{0,127,255} 
\definecolor{somecolor6}{RGB}{243,237,55} 
\definecolor{somecolor7}{RGB}{127,150,150} 
\definecolor{somecolor8}{RGB}{250,127,120} 
\definecolor{somecolor9}{RGB}{30,127,25} 

\begin{figure*}[t]
\begin{center}
\begin{mdframed}
\small

{(\textcolor{green}{I})\textsuperscript{\textbf{(1)}}}\textsubscript{P\_bather} [\textbf{\textcolor{orange}{decided}}]\textsubscript{E\_wash} to take a {(\textcolor{purp}{bath})\textsuperscript{\textbf{(2)}}}\textsubscript{P\_bath}  yesterday afternoon after working out .
Once {(\textcolor{green}{I})\textsuperscript{\textbf{(1)}}}\textsubscript{P\_bather} got back home , {(\textcolor{green}{I})\textsuperscript{\textbf{(1)}}}\textsubscript{P\_bather} [\textbf{\textcolor{orange}{walked}}]\textsubscript{E\_enter\_bathroom} to {(\textcolor{green}{my})\textsuperscript{\textbf{(1)}}}\textsubscript{P\_bather} {(\textcolor{somecolor}{bathroom})\textsuperscript{\textbf{(3)}}}\textsubscript{P\_bathroom} and first quickly scrubbed the {(\textcolor{somecolor2}{bathroom tub})\textsuperscript{\textbf{(4)}}}\textsubscript{P\_bathtub} by [\textbf{\textcolor{orange}{turning on}}]\textsubscript{E\_turn\_water\_on}  the {(\textcolor{somecolor3}{water})\textsuperscript{\textbf{(5)}}}\textsubscript{P\_water} and rinsing {(\textcolor{somecolor2}{it})\textsuperscript{\textbf{(4)}}}\textsubscript{P\_bathtub} clean with a rag .
After {(\textcolor{green}{I})\textsuperscript{\textbf{(1)}}}\textsubscript{P\_bather} finished , {(\textcolor{green}{I})\textsuperscript{\textbf{(1)}}}\textsubscript{P\_bather} [\textbf{\textcolor{orange}{plugged}}]\textsubscript{E\_close\_drain} the {(\textcolor{somecolor2}{tub})\textsuperscript{\textbf{(4)}}}\textsubscript{P\_bathtub} and began [\textbf{\textcolor{orange}{filling}}]\textsubscript{E\_fill\_water} {(\textcolor{somecolor2}{it})\textsuperscript{\textbf{(4)}}}\textsubscript{P\_bathtub} with warm {(\textcolor{somecolor3}{water})\textsuperscript{\textbf{(5)}}}\textsubscript{P\_water} set at about 98 {(\textcolor{somecolor5}{degrees})\textsuperscript{\textbf{(6)}}}\textsubscript{P\_temperature} .

\end{mdframed}
\caption{An excerpt from a story in the InScript corpus. The referring expressions are in parentheses, and the corresponding discourse referent label is given by the superscript. Referring expressions of the same discourse referent have the same color and superscript number. Script-relevant events are in square brackets and colored in orange. Event type is indicated by the corresponding subscript.}
\label{fig:example}
\end{center}
\end{figure*}

This paper represents foundational research investigating human language processing. However, it also has the potential for  application in assistant technology and embodied agents. The goal is to achieve human-level language comprehension in realistic settings, and in particular to achieve robustness in the face of errors or noise. Explicitly modelling expectations that are driven by common-sense knowledge is an important step in this direction.

In order to be able to investigate the influence of script knowledge on
discourse referent expectations, we use a corpus that
contains frequent reference to script knowledge, and provides  
annotations for coreference information, script events and participants (Section \ref{sec:corpus}).
In Section \ref{sec:ref-cloze}, we present a large-scale experiment for empirically assessing human expectations on upcoming referents, which allows us to quantify at what points in a text humans have very clear anticipations vs.~when they do not. Our goal is to model human expectations, even if they turn out to be incorrect  in a specific instance. The experiment was conducted via Mechanical Turk and follows the methodology of Tily and Piantadosi~\shortcite{tily2009refer}.
In section \ref{sec:drmodel}, we describe our computational model that represents script knowledge.
The model is trained on the gold standard annotations of the corpus, because we assume that human comprehenders usually will have an analysis of the preceding discourse which closely corresponds to the gold standard. We compare the prediction accuracy of this model to human predictions, as well as to two baseline models in Section \ref{sec:exp1}. One of them uses only
structural linguistic features for predicting referents; the other uses general script-independent
selectional preference features. 
In Section \ref{sec:refexppred}, we test whether surprisal (as estimated from human guesses vs.~computational models) can predict the type of referring expression used in the original texts in the corpus (pronoun vs.~full referring expression). This experiment also has wider implications with respect to the on-going discussion of whether the referring expression choice is dependent on predictability, as predicted by the uniform information density hypothesis.

The contributions of this paper consist of: 
\begin{itemize}[noitemsep,nolistsep]
\item a large dataset of human expectations, in a variety of texts related to every-day activities. 
\item an implementation of the conceptual distinction between the semantic level of referent prediction and the type of a referring expression.
\item a computational model which significantly improves modelling of human anticipations.
\item showing that script knowledge is a significant factor in human expectations.
\item testing the hypothesis of Tily and Piantadosi that the choice of the type of referring expression (pronoun or full NP) depends on the predictability of the referent.
 \end{itemize}



\subsection{Scripts}
\label{sec:scripts}
Scripts represent knowledge about typical event sequences
\cite{schank77}, for example the sequence of events happening
when eating at a restaurant. Script knowledge thereby includes events
like \textit{order}, \textit{bring} and \textit{eat} as well as
participants of those events, e.g., \textit{menu}, \textit{waiter},
\textit{food}, \textit{guest}. Existing methods for acquiring script
knowledge are based on extracting narrative chains from text
\cite{chambers2008unsupervised,chambers2009unsupervised,Jans2012,pichotta2014statistical,rudinger2015learning,modi2016,ahrendt2016}
or by eliciting script knowledge via Crowdsourcing on Mechanical Turk
\cite{Regneri2010,frermannhierarchical,modi2014}.

Modelling anticipated events and participants is motivated by evidence
showing that event representations in humans contain information not
only about the current event, but also about previous and future
states, that is, humans generate anticipations about event sequences
during normal language comprehension \cite{schutz2007prospective}.
Script knowledge representations have been shown to be useful in NLP
applications for ambiguity resolution during reference resolution
\cite{rahman2012resolving}. 



\section{Data: The InScript Corpus}
\label{sec:corpus}
Ordinary texts, including narratives, encode script structure in a way that is too complex and too implicit at the same time to enable a systematic study of script-based expectation. They contain interleaved references to many different scripts, and they usually refer to single scripts in a point-wise fashion only, relying on the ability of the reader to infer the full event chain using their background knowledge. 

We use the InScript corpus \cite{modiinscript} to study the predictive effect of script knowledge. 
InScript is a crowdsourced corpus of simple narrative texts.
Participants were asked to write about a specific activity (e.g., a restaurant visit, a bus
ride, or a grocery shopping event) which they personally
experienced, and they were instructed to tell the story as if explaining the
activity to a child. This resulted in stories that are centered around a specific scenario and that explicitly mention mundane
details. Thus, they generally realize longer event chains associated with a single script, which makes them particularly appropriate to our purpose.

The InScript corpus is labelled with event-type, participant-type, and coreference information. Full verbs are labeled with event type information, heads of all noun phrases with participant types, using scenario-specific lists of event types (such as \textit{enter  bathroom, close drain} and {\it fill water} for the ``taking a bath'' scenario) and participant types (such as \textit{bather, water} and {\it bathtub}). On average, each template offers a choice of 20 event types 
and 18 participant types. 
 
The InScript corpus consists of 910 stories addressing 10 scenarios (about 90 stories per scenario). The corpus has 200,000 words, 12,000 verb instances with event labels, and 44,000 head nouns with participant instances. Modi et al.~\shortcite{modiinscript} report an inter-annotator agreement of 0.64 for event types and 0.77 for participant types (Fleiss' kappa). 
 
We use gold-standard event- and participant-type annotation to study the influence of script knowledge on the expectation of discourse referents. In addition, InScript provides coreference annotation, which makes it possible to
keep track of the mentioned discourse referents at each point in the story. We use this information in the computational model of DR prediction and in the DR guessing experiment described in the next section. An example of an annotated InScript story is shown in Figure \ref{fig:example}.

\section{Referent Cloze Task}
\label{sec:ref-cloze}
\definecolor{orange}{RGB}{255,150,0}
\definecolor{purp}{RGB}{127,0,255}
\definecolor{somecolor}{RGB}{50,200,200} 
\definecolor{somecolor2}{RGB}{255,0,255} 
\definecolor{somecolor3}{RGB}{200,100,100} 
\definecolor{somecolor4}{RGB}{100,100,100} 
\definecolor{somecolor5}{RGB}{0,127,255} 
\definecolor{somecolor6}{RGB}{243,237,55} 
\definecolor{somecolor7}{RGB}{127,150,150} 
\definecolor{somecolor8}{RGB}{250,127,120} 
\definecolor{somecolor9}{RGB}{30,127,25} 
\definecolor{somecolor10}{RGB}{220,220,220} 

\begin{figure}[t]
\begin{center}
\mdfsetup{%
   middlelinecolor=black,
   middlelinewidth=1.5pt,
   backgroundcolor=white,
   roundcorner=20pt
}
\begin{mdframed}
\small

{(\textcolor{green}{I})\textsuperscript{\textbf{(1)}}} decided to take a {(\textcolor{purp}{bath})\textsuperscript{\textbf{(2)}}} yesterday afternoon after working out .
Once {(\textcolor{green}{I})\textsuperscript{\textbf{(1)}}} got back home , {(\textcolor{green}{I})\textsuperscript{\textbf{(1)}}} walked to {(\textcolor{green}{my})\textsuperscript{\textbf{(1)}}} {(\textcolor{somecolor}{bathroom})\textsuperscript{\textbf{(3)}}} and first quickly scrubbed the {(\textcolor{somecolor2}{bathroom tub})\textsuperscript{\textbf{(4)}}} by turning on the {(\textcolor{somecolor3}{water})\textsuperscript{\textbf{(5)}}} and rinsing {(\textcolor{somecolor2}{it})\textsuperscript{\textbf{(4)}}} clean with a rag .
After {(\textcolor{green}{I})\textsuperscript{\textbf{(1)}}} finished , {(\textcolor{green}{I})\textsuperscript{\textbf{(1)}}} plugged  \textcolor{white}{\colorbox{black}{\textbf{XXXXXX}}}
\end{mdframed}
\caption{An illustration of the Mechanical Turk experiment for the referent cloze task. Workers are supposed to guess the upcoming referent (indicated by \textcolor{white}{\colorbox{black}{\textbf{XXXXXX}}} above). They can either choose from the previously activated referents, or they can write something new. }
\label{fig:mturkExample}
\end{center}
\end{figure}


\begin{figure}[t]
\begin{center}
{\includegraphics[scale=0.15]{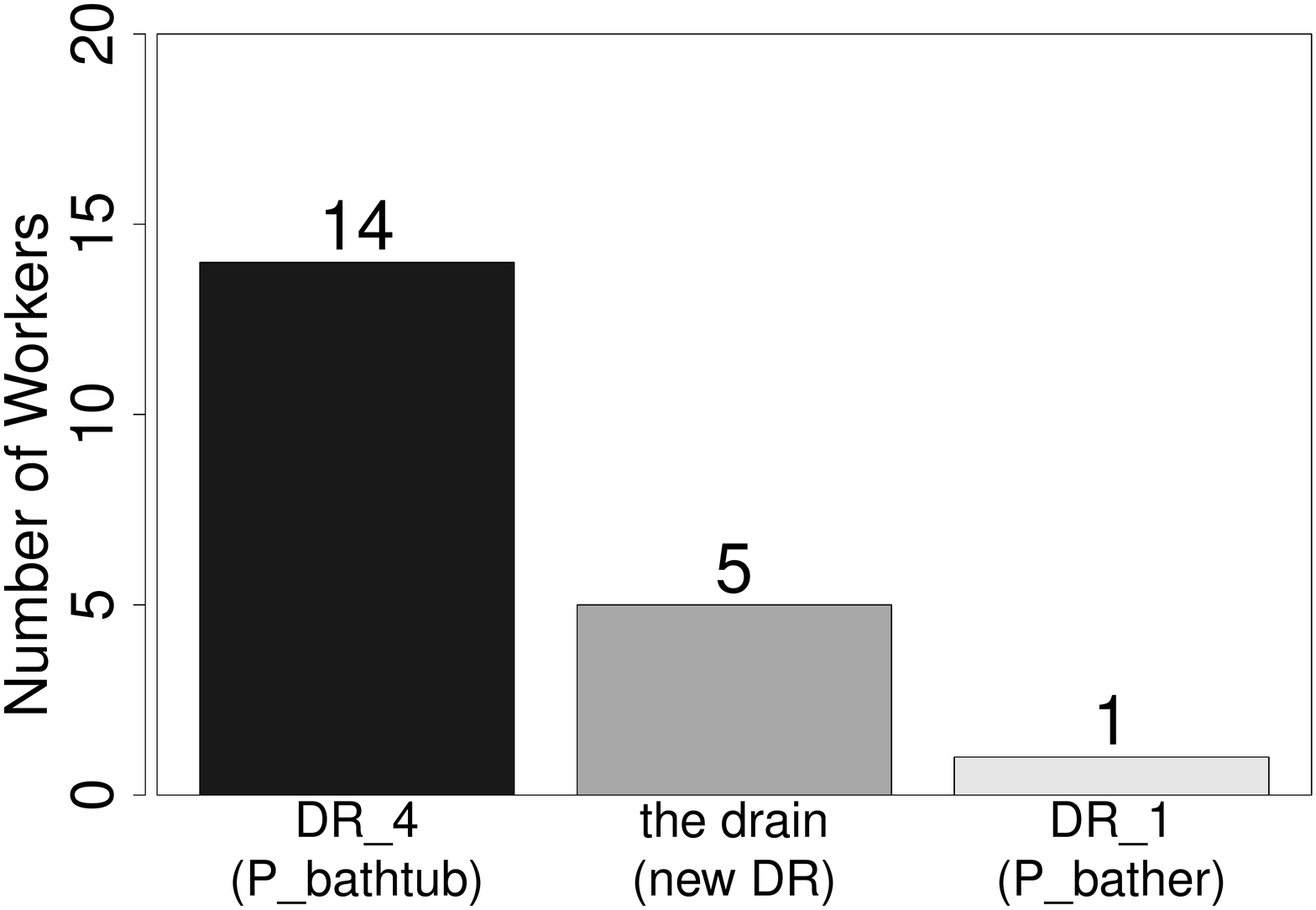}} 
\caption{Response of workers corresponding to the story in Fig.~\ref{fig:mturkExample}. Workers guessed two already activated discourse referents (DR) DR\_4 and DR\_1. Some of the workers also chose the ``new'' option and wrote different lexical variants of ``bathtub drain'', a new DR corresponding to the participant type ``the drain''.}
\label{fig:mturkResponse}
\end{center}
\end{figure}

We use the InScript corpus to develop computational models for the prediction of discourse referents (DRs) and to evaluate their prediction accuracy. This can be done by testing how often our models manage to reproduce the original discourse referent (cf.~also the ``narrative cloze'' task by \cite{chambers2008unsupervised} which tests whether a verb together with a role can be correctly guessed by a model). However, we do not only want to predict the ``correct'' DRs in a text but also to model human expectation of DRs in context. To empirically assess human expectation, we created an additional database of crowdsourced human predictions of discourse referents in context using Amazon Mechanical Turk. The design of our experiment closely resembles the guessing game of \cite{tily2009refer} but extends it in a substantial way.

Workers had to read stories of the InScript corpus \footnote{The corpus is available at : \url{http://www.sfb1102.uni-saarland.de/?page_id=2582}} and guess upcoming participants: for each target NP, workers were shown the story up to this NP excluding the NP itself, and they were asked to guess the next person or object most likely to be referred to. In case they decided in favour of a discourse referent already mentioned, they had to choose among the available discourse referents by clicking an NP in the preceding text, i.e., some noun with a specific, coreference-indicating color; see Figure \ref{fig:mturkExample}. Otherwise, they would click the ``New'' button, and  would in turn be asked to give a short description of the new person or object they expected to be mentioned. The percentage of guesses that agree with the actually referred entity was taken as a basis for estimating the surprisal. 

The experiment was done for all stories of the test set: 182 stories (20\%) of the InScript corpus, evenly taken from all scenarios. Since our focus is on the effect of script knowledge, we only considered those NPs as targets that are direct dependents of script-related events. Guessing started from the third sentence only in order to ensure that a minimum of context information was available. To keep the complexity of the context manageable, we restricted guessing to a maximum of 30 targets and skipped the rest of the story (this applied to 12\% of the stories).  We collected 20 guesses per NP for 3346 noun phrase instances, which amounts to a total of around 67K guesses. Workers selected a context NP in 68\% of cases and ``New" in 32\% of cases.

Our leading hypothesis is that script knowledge substantially influences human expectation of discourse referents. The guessing experiment provides a basis to estimate human expectation of already mentioned DRs (the number of clicks on the respective NPs in text). However, we expect that script knowledge has a particularly strong influence in the case of first mentions. Once a script is evoked in a text, we assume that the full script structure, including all participants, is activated and available to the reader. 

\newcite{tily2009refer} are interested in second mentions only and therefore do not make use of the worker-generated noun phrases classified as ``New". To study the effect of activated but not explicitly mentioned participants, we carried out a subsequent annotation step on the worker-generated noun phrases classified as ``New".  We presented annotators with these noun phrases in their contexts (with co-referring NPs marked by color, as in the M-Turk experiment) and, in addition, displayed all participant types of the relevant script (i.e., the script associated with the text in the InScript corpus). Annotators did not see the ``correct" target NP. We asked annotators to either (1) select the participant type instantiated by the NP (if any), (2) label the NP as unrelated to the script, or (3), link the NP to an overt antecedent in the text, in the case that the NP is actually a second mention that had been erroneously labeled as new by the worker.  
Option (1) provides a basis for a fine-grained estimation of first-mention DRs. Option (3), which we added when we noticed the considerable number of overlooked antecedents, serves as correction of the results of the M-Turk experiment. Out of the 22K annotated ``New" cases, 39\% were identified as second mentions, 55\% were linked to a participant type, and 6\% were classified as really novel.

\section{Referent Prediction Model}
\label{sec:drmodel}

In this section, we describe the model we use to predict upcoming
discourse referents (DRs). 
 
\subsection{Model}

Our model should not only assign probabilities to DRs already explicitly introduced in the preceding text fragment (e.g., ``bath'' or ``bathroom'' for the cloze task in Figure~\ref{fig:mturkExample})  but also reserve some probability mass for `new' DRs, i.e., DRs activated via the script context or completely novel ones not belonging to the script. In principle, different variants of the activation mechanism must be distinguished. For many participant types, a single participant belonging to a specific semantic class is expected (referred to with \textit{the bathtub} or \textit{the soap}). In contrast, the ``towel'' participant type may activate a set of objects, elements of which then can be referred to with  \textit{a towel} or \textit{another towel}. The ``bath means'' participant type may even activate a group of DRs belonging to different semantic classes (e.g., {\it bubble bath} and {\it salts}). Since it is not feasible to enumerate all potential participants, for `new' DRs we only predict their participant type (``bath means'' in our example).  In other words, the number of categories in our model  is equal to the number of previously introduced DRs plus the number of participant types of the script plus 1, reserved for a new DR not corresponding to any script participant (e.g., {\it cellphone}). In what follows, we slightly abuse the terminology and refer to all these categories as discourse referents.

Unlike standard co-reference models, which predict co-reference chains relying on the entire document, our model is  incremental, that is,  when predicting a discourse referent $d^{(t)}$ at a given position $t$,
it can look only in the history $h^{(t)}$ (i.e., the preceding part of the document), excluding the referring expression (RE) for the predicted DR. We also assume that past REs are correctly resolved and assigned to correct participant types (PTs). 
Typical NLP applications use automatic coreference resolution systems, but since we want to model human behavior, this might be inappropriate, since an automated system would underestimate human performance. This may be a strong assumption, but for reasons explained above, we use gold standard past REs.


We use the following log-linear model (``softmax regression''):
\vspace{-0.2in}
\par\noindent
\resizebox{\columnwidth}{!}{
  \begin{minipage}{\columnwidth}
\begin{align*}
\tiny
p(d^{(t)} = d | h^{(t)}) = \frac{
\exp(\mathbf{w}^{T} \mathbf{f}(d, h^{(t)} ))}
{ \sum_{d'}{\exp(\mathbf{w}^{T} \mathbf{f}(d', h^{(t)} ))}
},
\end{align*}
\end{minipage}
}
\normalsize

\vspace{0.1in} 
where $\mathbf{f}$ is the feature function we will discuss in the following subsection, $\mathbf{w}$ are model parameters, and
the summation in the denominator is over the set of categories described above.


Some of the features included in $\mathbf{f}$ are a function of the predicate syntactically governing the unobservable target RE (corresponding to the DR being predicted). However, in our incremental setting, the predicate is not available in the history $h^{(t)}$ for subject NPs. In this case, we use an additional probabilistic model, which estimates the probability of the predicate $v$ given the context $h^{(t)}$, and marginalize out its predictions:
\resizebox{1\columnwidth}{!}{
  \begin{minipage}{\columnwidth}
\begin{align}
\tiny
p( & d^{(t)}\!=\! d | h^{(t)}) \!=\!\! \mathlarger{\mathlarger{\sum}_{v}} { 
p(v | h^{(t)})
\frac{\exp(\mathbf{w}^{T} \mathbf{f}(d, h^{(t)}, v ))}
{ \sum_{d'}{\exp(\mathbf{w}^{T} \mathbf{f}(d', h^{(t)}, v ))}}
}\!\!
\nonumber
\end{align}
\end{minipage}
}
\normalsize
\\
The predicate probabilities $p(v | h^{(t)})$ are computed based on the sequence of preceding predicates (i.e., ignoring any other words) using the recurrent 
neural network language model estimated on our training set.\footnote{We used RNNLM toolkit  \cite{mikolov2011rnnlm,mikolov2010recurrent} with default settings.}
The expression $ \mathbf{f}(d, h^{(t)}, v )$ denotes the feature function computed for the referent $d$, given the history composed of $h^{(t)}$ and the predicate $v$.


\begin{table}[bt]
\begin{center}
\small
\begin{tabular}{|c|c|}
\hline
Feature & Type \\
\hline
\hline
Recency & Shallow Linguistic \\
\hline
Frequency & Shallow Linguistic \\
\hline
Grammatical function & Shallow Linguistic \\
\hline
Previous subject & Shallow Linguistic \\
\hline
Previous object & Shallow Linguistic \\
\hline
Previous RE type & Shallow Linguistic \\
\hline
Selectional preferences & Linguistic \\
\hline
Participant type fit & Script \\
\hline
Predicate schemas & Script  \\
\hline
\end{tabular}
\end{center}
\caption{Summary of feature types}
\label{table:featuretable}
\end{table}%

\subsection{Features}

Our features encode properties of a DR as well as characterize its compatibility with the context. 
We face two challenges when designing our features. First, although the sizes of our datasets are respectable from the script annotation perspective, they are too small to learn a richly parameterized model.  For many of our features, we address this challenge by using external word embeddings\footnote{We use 300-dimensional word embeddings estimated on Wikipedia with the skip-gram model of Mikolov et al. \shortcite{mikolov2013distributed}: \url{https://code.google.com/p/word2vec/}} and associate parameters with some simple similarity measures computed using these embeddings. Consequently, there are only a few dozen parameters which need to be estimated from scenario-specific data. Second, in order to test our hypothesis that script information is beneficial for the DR prediction task, we need to disentangle the influence of script information from general linguistic knowledge. We address this by carefully splitting the features apart, even if it prevents us from modeling some interplay between the sources of information. We will describe both classes of features below; also see a summary in Table~\ref{table:featuretable}.



\subsubsection{ Shallow Linguistic Features}

These features are based on Tily and Piantadosi~\shortcite{tily2009refer}. In addition, we consider a selectional preference feature. 

\noindent \textbf{Recency feature.} This feature captures the distance $l_t(d)$ between the position $t$ and the last occurrence of the candidate
DR $d$. As a distance measure, we use the number of sentences from the last mention and exponentiate this number to make the dependence more extreme; only very recent DRs will receive a noticeable weight: $\exp(-l_t(d))$.  This feature is set to $0$ for new DRs.

\noindent \textbf{Frequency.} The frequency feature indicates the number of times the candidate discourse referent $d$ has been mentioned so far. We do not perform any bucketing.

\noindent \textbf{Grammatical function.} This feature encodes the dependency relation assigned to the head word of the last mention of the DR or a special
{\tt none} label if the DR is new.

\noindent \textbf{Previous subject indicator.} This binary feature indicates whether the candidate DR $d$ is coreferential with the subject of the  previous verbal predicate.

\noindent \textbf{Previous object indicator.} The same but for the object position.

\noindent \textbf{Previous RE type.} This three-valued feature indicates whether the previous mention of the candidate DR $d$ is a pronoun, a non-pronominal noun phrase, or has never been observed before.

\subsubsection{ Selectional Preferences Feature}
The selectional preference feature captures how well the candidate DR $d$ fits a given syntactic position $r$ of a given verbal predicate $v$. It is computed as the cosine similarity  $\mathrm{sim_{cos}}(\mathbf{x}_d^T, \mathbf{x}_{v,r})$  of a vector-space representation of the DR $\mathbf{x}_d$   
and a structured vector-space representation of the predicate $\mathbf{x}_{v,r}$. The similarities are calculated using a Distributional Memory approach similar to that of Baroni and Lenci~\shortcite{baroni2010distributional}. Their structured vector space representation has been shown to work well on tasks that evaluate correlation with human thematic fit estimates \cite{baroni2010distributional,baroni2014,sayeedrepeval2016} and is thus suited to our task. 

The representation $\mathbf{x}_d$ is computed as an average of head word representations of all the previous mentions of DR $d$, where the word vectors are obtained from the TypeDM model of Baroni and Lenci~\shortcite{baroni2010distributional}. This is a count-based, third-order co-occurrence tensor whose indices 
are a word $w_0$, a second word $w_1$, and a complex syntactic relation $r$, which is used as a stand-in for a semantic link.  The values for each $(w_0,r,w_1)$ cell
of the tensor are the local mutual information (LMI) estimates obtained from 
a dependency-parsed combination of large corpora (ukWaC, BNC, and Wikipedia).

Our procedure has some differences with that of Baroni and Lenci. For
example, for estimating the fit of an alternative new DR (in other
words, $\mathbf{x}_d$ based on no previous mentions), we use an
average over head words of all REs in the training set, a ``null
referent.''  $\mathbf{x}_{v,r}$ is calculated as the average of the
top 20 (by LMI) $r$-fillers for $v$ in TypeDM; in other words, the
prototypical instrument of {\it rub} may be represented by summing
vectors like {\it towel, soap, eraser, coin\ldots} If the predicate
has not yet been encountered (as for subject positions), scores for
all scenario-relevant verbs are emitted for marginalization.

\subsubsection{Script Features}

In this section, we describe features which rely on  script information. Our goal will be to show that such common-sense
information is beneficial in performing DR prediction. We consider only two script features. 




\vspace{+1ex}
\noindent \textbf{Participant type fit} 

\noindent This feature characterizes how well the participant type (PT) of the candidate DR $d$ fits a specific syntactic role $r$ of the governing predicate $v$; it can be regarded as a generalization of the selectional preference feature to participant types and also its specialisation to the considered scenario. Given the candidate DR $d$, its participant type $p$, and the syntactic relation $r$, we collect all the predicates in the training set which have the participant type $p$ in the position~$r$. The embedding of the DR $\mathbf{x}_{p, r}$ is given by the average embedding of these predicates. The feature is computed as the dot product of $\mathbf{x}_{p, r}$ and the word embedding of the predicate $v$.

\vspace{+1ex}
\noindent \textbf{Predicate schemas}

\definecolor{orange}{RGB}{255,150,0}
\definecolor{purp}{RGB}{127,0,255}
\definecolor{somecolor}{RGB}{50,200,200} 
\definecolor{somecolor2}{RGB}{255,0,255} 
\definecolor{somecolor3}{RGB}{200,100,100} 
\definecolor{somecolor4}{RGB}{100,100,100} 
\definecolor{somecolor5}{RGB}{0,127,255} 
\definecolor{somecolor6}{RGB}{243,237,55} 
\definecolor{somecolor7}{RGB}{127,150,150} 
\definecolor{somecolor8}{RGB}{250,127,120} 
\definecolor{somecolor9}{RGB}{30,127,25} 
\definecolor{somecolor10}{RGB}{220,220,220} 

\begin{figure}[t]
\begin{center}
\mdfsetup{%
   middlelinecolor=black,
   middlelinewidth=1.5pt,
   backgroundcolor=white,
   roundcorner=20pt
}
\begin{mdframed}
\small

{(\textcolor{green}{I})\textsuperscript{\textbf{(1)}}} decided to take a {(\textcolor{purp}{bath})\textsuperscript{\textbf{(2)}}} yesterday afternoon after working out .
{(\textcolor{green}{I})\textsuperscript{\textbf{(1)}}} was getting ready to go out and needed to get cleaned before {(\textcolor{green}{I})\textsuperscript{\textbf{(1)}}} went so {(\textcolor{green}{I})\textsuperscript{\textbf{(1)}}} decided to take a {(\textcolor{purp}{bath})\textsuperscript{\textbf{(2)}}}.  
 {(\textcolor{green}{I})\textsuperscript{\textbf{(1)}}} filled the {(\textcolor{somecolor2}{bathtub})\textsuperscript{\textbf{(3)}}} with warm  {(\textcolor{somecolor3}{water})\textsuperscript{\textbf{(4)}}} and added some {(\textcolor{somecolor8}{bubble bath})\textsuperscript{\textbf{(5)}}}. 
 {(\textcolor{green}{I})\textsuperscript{\textbf{(1)}}} got undressed and stepped into the {(\textcolor{somecolor3}{water})\textsuperscript{\textbf{(4)}}}.  
 {(\textcolor{green}{I})\textsuperscript{\textbf{(1)}}} grabbed the {(\textcolor{somecolor5}{soap})\textsuperscript{\textbf{(5)}}} and rubbed it on {(\textcolor{green}{my})\textsuperscript{\textbf{(1)}}} {(\textcolor{somecolor4}{body})\textsuperscript{\textbf{(7)}}} and rinsed \textcolor{white}{\colorbox{black}{\textbf{XXXXXX}}}
\end{mdframed}
\caption{An example of the referent cloze task. Similar to the Mechanical Turk experiment (Figure \ref{fig:mturkExample}), our referent prediction model is asked to guess the upcoming DR.}
\label{fig:mturkExample2}
\end{center}
\end{figure}

%

\noindent The following feature captures a specific aspect of knowledge about prototypical sequences of events. This knowledge is called {\it predicate schemas} in the recent co-reference modeling work of Peng et al.~\shortcite{peng2015}.  In predicate schemas, the goal is to model pairs
of events such that if a DR $d$ participated in the first event (in a specific role), it is likely to participate in the second event (again, in a specific role).  
For example, in the restaurant scenario, if one observes a phrase {\it John ordered}, one is likely to see {\it John waited} somewhere later in the document.  Specific arguments are not that important (where it is {\it John} or some other DR), what is important is that the argument is reused across the predicates. This would correspond to the rule {\it X-subject-of-order} $\rightarrow$ {\it X-subject-of-eat}.\footnote{In this work,
we limit ourselves to rules where the syntactic function is the same on both sides of the rule. In other words, we can, in principle, encode the pattern
{\it X pushed Y} $\rightarrow$ {\it X apologized} but not the pattern 
{\it X pushed Y} $\rightarrow$ {\it Y cried}.}
Unlike the previous work, our dataset is small, so we cannot induce these rules directly as there will be very few rules, and the model would not generalize to new data well enough.  Instead, we again encode this intuition using similarities in the real-valued embedding space.

Recall that our goal is to compute a feature $\varphi(d, h^{(t)})$ indicating how likely a potential DR $d$ is to follow, given the history $h^{(t)}$.
For example,  imagine that the model is asked to predict the DR marked by XXXXXX in Figure~\ref{fig:mturkExample2}.  Predicate-schema rules can only yield previously introduced DRs, so the score $\varphi(d, h^{(t)}) = 0$ for any new DR $d$. 
Let us use ``soap'' as an example of a previously introduced DR and see how the feature is computed. In order to choose which inference rules can be applied to yield ``soap'', we can inspect
Figure \ref{fig:mturkExample2}. There are only two preceding predicates which have DR ``soap'' as their object ({\it rubbed} and {\it grabbed}), resulting in two potential
 rules {\it X-object-of-grabbed} $\rightarrow$ {\it X-object-of-rinsed}  and  {\it X-object-of-rubbed} $\rightarrow$ {\it X-object-of-rinsed}.
We define the score $\varphi(d, h^{(t)})$ as the average of the rule scores.
More formally, we can write
\begin{align}
\label{eq:phi1}
\varphi(d, h^{(t)})\! =\! 
\frac{1}{|N(d, h^{(t)})|} 
\! \sum_{(u,v,r) \in  N(d, h^{(t)})}{ \!\!\!\!\!\!\!\! \!\!\!\!  \psi(u, v, r)},
\end{align}
where $\psi(u, v, r)$ is the score for a rule {\it X-r-of-u} $\rightarrow$ {\it X-r-of-v},  
$N(d, h^{(t)})$ is the set of applicable rules,  and $| N(d, h^{(t)}) |$ denotes its cardinality.\footnote{In all our experiments, rather than considering all potential predicates in the history to instantiate rules,
we take into account only  2 preceding verbs. In other words, $u$ and $v$ can be interleaved by at most one verb and $|N(d, h^{(t)})|$ is in  $\{0,1,2\}$.}  We define $\varphi(d, h^{(t)})$ as $0$, when the set of applicable rules is empty (i.e. $| N(d, h^{(t)}) | = 0$).

\begin{table}[bt]
\begin{center}
\Large
\resizebox{\columnwidth}{!}{%
\begin{tabular}{|P{25mm}|P{25mm}|P{60mm}|}
\hline
\textbf{Model Name} & \textbf{Feature Types} & \textbf{Features}  \\ \hline
\textbf{Base} & Shallow Linguistic Features &  Recency, Frequency, \newline Grammatical function, \newline Previous subject, \newline Previous object \hphantom{w} \hphantom{w}
 \\ \hline
\textbf{Linguistic} & Shallow Linguistic \hphantom{w}Features \newline \hphantom{w} \textbf{+}  \newline Linguistic Feature  &  Recency, Frequency, \newline Grammatical function, \newline Previous subject, \newline  Previous object \newline \hphantom{w} \textbf{+} \newline  Selectional Preferences  \hphantom{w}
\\ \hline
\textbf{Script} &  Shallow Linguistic \hphantom{w}Features \newline  \hphantom{w} \textbf{+} \newline Linguistic \hphantom{w}Feature  \newline \hphantom{w} \textbf{+} \newline Script Features &  Recency, Frequency, \newline Grammatical function, \newline Previous subject, \newline  Previous object \newline \hphantom{w} \textbf{+} \newline  Selectional Preferences \newline \hphantom{w} \textbf{+} \newline Participant type fit,\newline Predicate schemas\hphantom{w}\hphantom{w}
\\ \hline
\end{tabular}
}
\end{center}
\caption{Summary of model features}
\label{table:modelfeaturetable}
\end{table}%

 The scoring function $\psi(u, v, r)$ as a linear function of 
a joint embedding $\mathbf{x}_{u,v}$ of verbs $u$ and $v$:
\begin{align}
\nonumber
\psi(u, v, r) = \mathbf{\alpha}_r^T \mathbf{x}_{u,v}.
\end{align}
The two remaining questions are (1) how to define the joint embeddings $\mathbf{x}_{u,v}$, and (2) how to estimate the parameter
vector $\mathbf{\alpha}_r$.  The joint embedding of two predicates, $\mathbf{x}_{u,v}$, can, in principle, be any composition function of 
embeddings of $u$ and $v$, for example their sum or component-wise product. Inspired by Bordes et al.~\shortcite{bordes2013translating}, we use the difference between the word embeddings:
\begin{align}
\nonumber
\psi(u, v, r) = \mathbf{\alpha}_r^T (\mathbf{x}_u - \mathbf{x}_v),
\end{align}
where $\mathbf{x}_u$ and  $\mathbf{x}_v$ are external embeddings of the corresponding verbs. 
Encoding the succession relation as translation in the embedding space has one desirable property:
the scoring function will be largely agnostic to the morphological form of the predicates. For example, the difference between the embeddings of $rinsed$ and $rubbed$ is very similar to that of $rinse$ and $rub$~\cite{botha2014compositional}, so the corresponding rules will receive similar scores. Now, we can rewrite the equation~(\ref{eq:phi1}) as
\begin{align}
\varphi(d, h^{(t)})\!  = 
\mathbf{\alpha}_{r(h^{(t)})}^T \!\!
\frac{  \sum_{(u,v,r) \in  N(d, h^{(t)})}{ (\mathbf{x}_u - \mathbf{x}_v) }}{| N(d, h^{(t)})|}  \label{eq:phi2}
\end{align}
where $r(h^{(t)})$ denotes the syntactic function corresponding to the DR being predicted (object in our example).
 

\begin{table*}[ht]
\centering
\large
\resizebox{2\columnwidth}{!}{%
\begin{tabular}{|c||c|c||c|c||c|c||c|c||}
\hline
Scenario                      & \multicolumn{2}{c||}{Human Model} & \multicolumn{2}{c||}{Script Model} & \multicolumn{2}{c||}{Linguistic Model} & \multicolumn{2}{c||}{Tily Model} \\ \hline
                              & Accuracy       & Perplexity      & Accuracy       & Perplexity       & Accuracy         & Perplexity         & Accuracy      & Perplexity      \\ \hline
Grocery Shopping              & 74.80          & 2.13            & 68.17          & 3.16             & 53.85            & 6.54               & 32.89         & 24.48           \\ \hline
Repairing a flat bicycle tyre & 78.34          & 2.72            & 62.09          & 3.89             & 51.26            & 6.38               & 29.24         & 19.08           \\ \hline
Riding a public bus           & 72.19          & 2.28            & 64.57          & 3.67             & 52.65            & 6.34               & 32.78         & 23.39           \\ \hline
Getting a haircut             & 71.06          & 2.45            & 58.82          & 3.79             & 42.82            & 7.11               & 28.70         & 15.40           \\ \hline
Planting a tree               & 71.86          & 2.46            & 59.32          & 4.25             & 47.80            & 7.31               & 28.14         & 24.28           \\ \hline
Borrowing book from library   & 77.49          & 1.93            & 64.07          & 3.55             & 43.29            & 8.40               & 33.33         & 20.26           \\ \hline
Taking Bath                   & 81.29          & 1.84            & 67.42          & 3.14             & 61.29            & 4.33               & 43.23         & 16.33           \\ \hline
Going on a train              & 70.79          & 2.39            & 58.73          & 4.20             & 47.62            & 7.68               & 30.16         & 35.11           \\ \hline
Baking a cake                 & 76.43          & 2.16            & 61.79          & 5.11             & 46.40            & 9.16               & 24.07         & 23.67           \\ \hline
Flying in an airplane         & 62.04          & 3.08            & 61.31          & 4.01             & 48.18            & 7.27               & 30.90         & 30.18           \\ \hline \hline
Average                       & 73.63          & 2.34            & 62.63          & 3.88             & 49.52            & 7.05               & 31.34         & 23.22           \\ \hline
\end{tabular}%
}
\caption{Accuracies (in \%) and perplexities for different models and scenarios. The script model substantially outperforms linguistic and base models (with $p < 0.001$, significance tested with McNemar's test \protect\cite{everitt1992analysis}). As expected, the human prediction model outperforms the script model (with $p < 0.001$, significance tested by McNemar's test). }
\label{table:accuracy}
\end{table*}

\begin{table}[h]
\centering
\Large
\resizebox{0.9\columnwidth}{!}{%
\begin{tabular}{|c|c|c|}
\hline
 Model & Accuracy & Perplexity \\ \hline
 Linguistic Model            & 49.52 &   7.05 \\ \hline
Linguistic Model +  Predicate Schemas  & 55.44 &  5.88 \\ \hline
Linguistic Model +  Participant type fit    & 58.88  & 4.29 \\ \hline
Full Script Model (both features)         & 62.63  &  3.88 \\ \hline
\end{tabular}%
}
\caption{Accuracies from ablation experiments.}
\label{table:ablation-accuracy}
\end{table}

 
As for the parameter vector $\mathbf{\alpha}_r$, there are again a number of potential ways how it can be estimated. For example, one can train a discriminative classifier to estimate the parameters. However, we opted for a simpler approach---we set it equal to the empirical estimate of the expected feature vector $x_{u,v}$ on the training set:\footnote{This essentially corresponds to using the Naive Bayes model with the simplistic assumption
that the score differences are normally distributed with spherical covariance matrices.}
\begin{align}
\label{eq:alpha}
\mathbf{\alpha}_r =  {1 \over D_r} \sum_{l,t}  \delta_r(r(h^{(l, t)})) \!\!\!\!\!\!\!\! \!\!\!\!\!\!\!\!   \sum_{(u,v,r') \in  N(d^{(l,t)}, h^{(l,t)})}{ \!\!\!\!\!\!\!\! \!\!\!\!\!\!\!\!  ( \mathbf{x}_u - \mathbf{x}_v)  },
\end{align}
where $l$ refers to a document in the training set, $t$ is (as before) a position in the document,
$h^{(l,t)}$ and $d^{(l,t)}$ are the history and the correct DR for this position, respectively.
The term  $\delta_r(r')$ is the Kronecker delta which equals $1$ if $r = r'$ and $0$, otherwise. $D_r$ is the total number 
of rules for the syntactic function $r$ in the training set:
\begin{align}
\nonumber
D_r =  \sum_{l,t} \delta_r(r(h^{(l,t)}))  \times  |N(d^{(l,t)}, h^{(l,t)}) |.
\end{align}
Let us illustrate the computation with an example.
Imagine that our training set consists of the document in Figure 1, and the trained model is used to predict the upcoming DR in our referent cloze example (Figure \ref{fig:mturkExample2}). The training document includes
the pair {\it X-object-of-scrubbed} $\rightarrow$ {\it X-object-of-rinsing}, so the corresponding term
($\mathbf{x}_\text{\it scrubbed}$ - $\mathbf{x}_\text{\it rinsing}$)
participates in the summation~(\ref{eq:alpha}) for $\mathbf{\alpha}_{obj}$.
As we rely on external embeddings, which encode semantic similarities between lexical items, the dot product
of this term and ($\mathbf{x}_\text{\it rubbed}$ - $\mathbf{x}_\text{\it rinsed}$) will be high.\footnote{The score would have been even  higher, should the predicate  be in the morphological form {\it rinsing} rather than {\it rinsed}. However, embeddings of {\it rinsing} and {\it rinsed} would still be sufficiently close to each other for our argument to hold.} Consequently, $\varphi(d, h^{(t)})$ is expected to be positive for $d = \text{``soap''}$, thus, predicting ``soap'' as the likely forthcoming DR. Unfortunately, there are other terms $(\bf{x}_u - \bf{x}_v)$ both in expression~(\ref{eq:alpha}) for $\mathbf{\alpha}_{obj}$ and in expression~(\ref{eq:phi2}) for $\varphi(d, h^{(t)})$. These terms may be irrelevant to the current prediction, as 
{\it X-object-of-plugged} $\rightarrow$ {\it X-object-of-filling} from Figure 1, and may not even encode any valid regularities, as
{\it X-object-of-got} $\rightarrow$ {\it X-object-of-scrubbed} (again from Figure 1). This may suggest that our feature will be too contaminated with noise to be informative for making predictions. However, recall that
independent random vectors in high dimensions are almost orthogonal, and, assuming they are bounded, their dot products are close to zero. Consequently, 
the products of the relevant (``non-random'') terms, in our example ($\mathbf{x}_\text{\it scrubbed}$ - $\mathbf{x}_\text{\it rinsing}$) and ($\mathbf{x}_\text{\it rubbed}$ - $\mathbf{x}_\text{\it rinsed}$),   are likely to overcome the (``random'') noise. As we will see in the ablation studies, the predicate-schema feature is indeed predictive of a DR and contributes
 to the performance of the full model.

\subsection{Experiments}
\label{sec:exp1}
We would like to test whether our 
model can produce accurate predictions and whether the model's guesses correlate well with human
predictions for the referent cloze task.

In order to be able to evaluate the effect of script knowledge on
referent predictability, we compare three models: our full
\textit{Script model} uses all of the features introduced in section
4.2; the \textit{Linguistic model} relies only on the `linguistic
features' but not the script-specific ones; and the \textit{Base model} includes all the shallow linguistic features. The Base model differs from the linguistic model in that it does not model selectional preferences. Table \ref{table:modelfeaturetable} summarizes features used in different models.


The data set was randomly divided into training (70\%), development (10\%, 91
stories from 10 scenarios), and test (20\%, 182 stories from 10 scenarios) sets.
The feature weights were learned using L-BFGS \cite{byrd1995limited} to optimize the log-likelihood.




\noindent\textbf{Evaluation against original referents.} We calculated the percentage of correct DR predictions.  See Table \ref{table:accuracy} for
the averages across 10 scenarios. 
We can see that the task appears hard for humans: their average performance reaches only 73\% accuracy.
 As expected, the Base model is the weakest system (the accuracy of 31\%). Modeling selectional preferences yields an extra 18\% in accuracy (Linguistic model).
The key finding is that incorporation of script knowledge increases the accuracy by further 13\%, although still far behind human performance (62\% vs. 73\%). 
Besides accuracy, we use perplexity, which we computed not only for all our models but also for human predictions. This was possible as each
task was solved by multiple humans. We used unsmoothed normalized guess frequencies as the probabilities.  
 As we can see from Table \ref{table:accuracy}, the perplexity scores are consistent with the accuracies: the script model again outperforms other methods, and, as expected, all the models are weaker than humans.
 
As we used two sets of script features, capturing different aspects of script knowledge, we performed extra ablation studies (Table \ref{table:ablation-accuracy}). The experiments confirm that both feature sets were beneficial.

\noindent\textbf{Evaluation against human expectations.}  In the previous subsection, we demonstrated that the incorporation of selectional preferences and, perhaps more interestingly, the integration of automatically acquired script knowledge lead to improved accuracy in predicting discourse referents. Now we turn to another question raised in the introduction: does incorporation of this knowledge make our predictions more human-like? In other words, are we able to accurately estimate human expectations? This includes not only being sufficiently accurate but also 
making the same kind of incorrect predictions.

\begin{figure}
\centering
{\includegraphics[scale=0.15, trim=0cm 1.1cm 0cm 0cm, clip=TRUE]{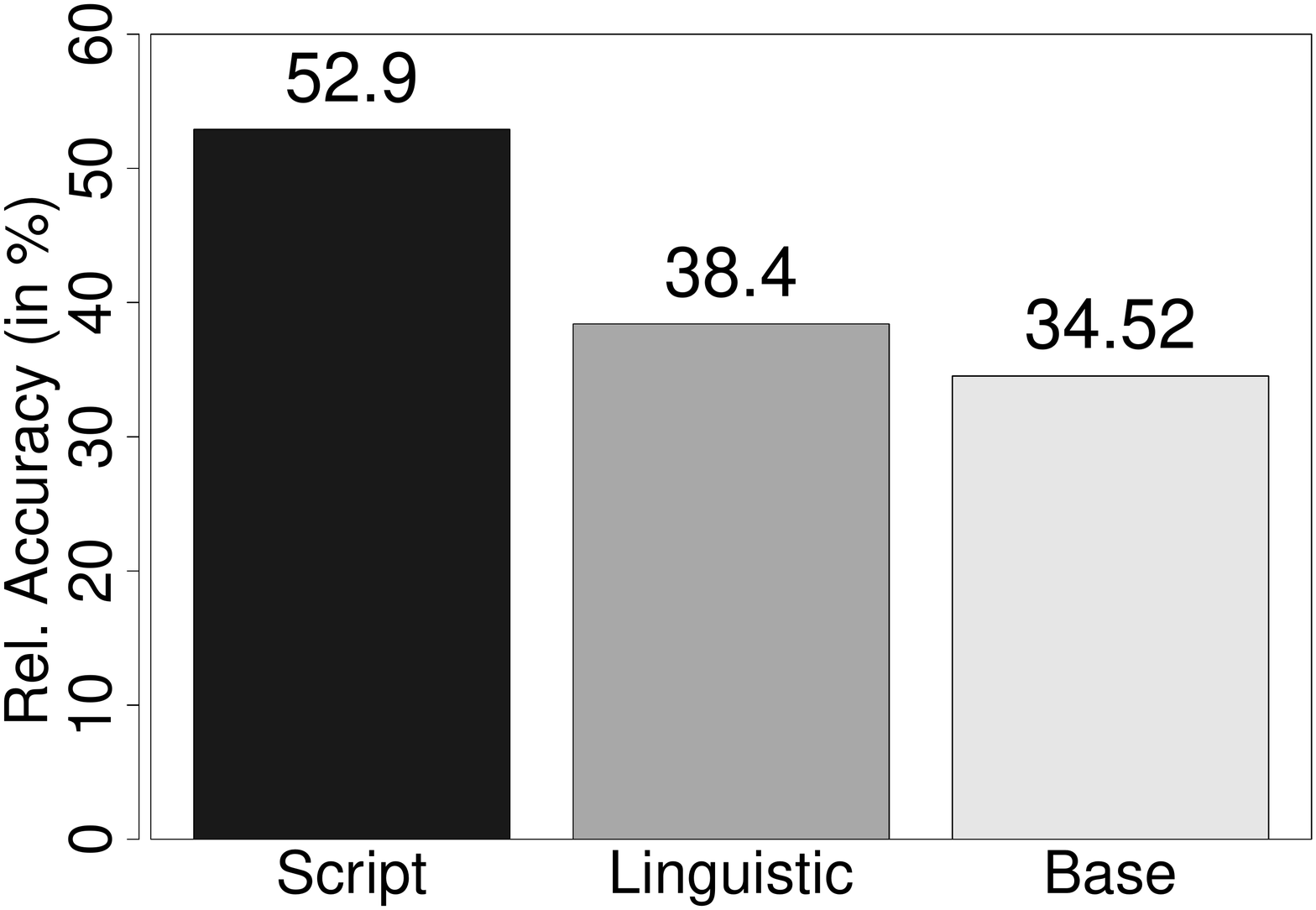}} 
\caption{Average relative accuracies of different models w.r.t human predictions.}
\label{fig:relAccuracy} 
\end{figure}

\begin{figure}
\centering
{\includegraphics[scale=0.15, trim=0cm 1.1cm 0cm 0cm, clip=TRUE]{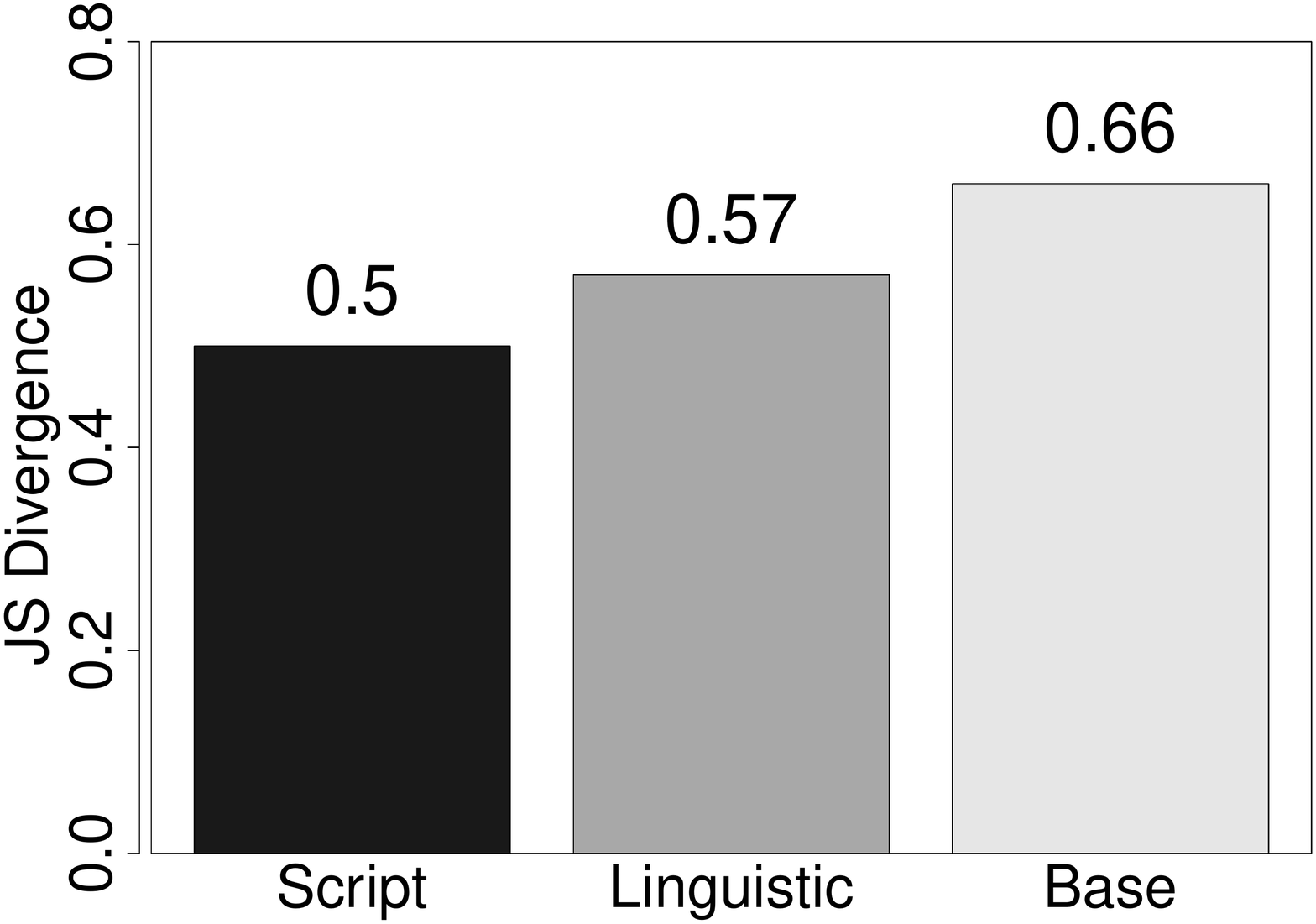}} 
\caption{Average Jensen-Shannon divergence between human predictions and  models.}
\label{fig:jsDist}
\end{figure}

In this evaluation, we therefore use human guesses collected during
the referent cloze task as our target. We then calculate the relative
accuracy of each computational model. As can be seen in Figure
\ref{fig:relAccuracy}, the Script model, at approx.~53\% accuracy, is a
lot more accurate in predicting human guesses than the Linguistic
model and the Base model. We can also observe that the margin
between the Script model and the Linguistic model is a lot
larger in this evaluation than between the Base model and the
Linguistic model. This indicates that the model which
has access to script knowledge is much more similar to human
prediction behavior in terms of top guesses than the script-agnostic models.

Now we would like to assess if our predictions are similar as distributions rather
than only yielding similar top predictions. In order to compare the distributions,
we use the
Jensen-Shannon divergence (JSD),
a symmetrized version of the Kullback-Leibler divergence.  

Intuitively, JSD measures the distance between two probability distributions. A smaller JSD value is indicative of more similar distributions.
Figure~\ref{fig:jsDist} shows that the probability
distributions resulting from the Script model are more similar to human
predictions than those of the Linguistic and Base models.

In these experiments, we have shown that 
script knowledge improves predictions of upcoming referents and that
the script model is the best among our models in approximating human referent predictions. 


\section{Referring Expression Type Prediction Model (RE Model)}
\label{sec:refexppred}
Using the referent prediction models, we next attempt to replicate Tily and Piantadosi's findings that the choice of the type of referring expression (pronoun or full NP) depends in part on the predictability of the referent. 

\subsection{Uniform Information Density hypothesis}
\label{sec:uid}

The uniform information density (UID) hypothesis
suggests that speakers tend to convey information at a uniform rate \cite{jaeger2010redundancy}.
Applied to choice of referring expression type, it would predict
that a highly predictable referent should be encoded using a short code (here: a pronoun), while an unpredictable referent should be encoded using a longer form (here: a full NP). 
Information density is measured using the
information-theoretic measure of the \textit{surprisal} $S$ of a message $m_i$:  
\vspace{-3pt}
\begin{equation*}
\small
 S(m_i) = - \log P(m_i \mid context)
 \label{eq:surprisal}
\end{equation*}
\normalsize

\vspace{-3pt}

UID has been very successful in explaining a variety of linguistic phenomena; see \newcite{jaeger2016signal}.  There is, however, controversy about whether UID affects pronominalization.
 \newcite{tily2009refer} report evidence that writers are more likely to refer using a pronoun or
proper name when the referent is easy to guess and use a full NP when
readers have less certainty about the upcoming referent; see also
\newcite{arnold2001}. But other experiments (using highly controlled stimuli)
have failed to find an effect of predictability on pronominalization
\cite{stevenson1994thematic,fukumura2010choosing,rohde2014grammatical}.
The present study hence contributes to the debate on whether UID
affects referring expression choice.

\begin{table*}[htp]
\centering
\Large
\resizebox{2\columnwidth}{!}{%
\begin{tabular}{|c|c|c|c|c|c|c|c|c|c|c|c|c|}
\hline
                & \multicolumn{4}{c|}{Estimate}         & \multicolumn{4}{c|}{Std. Error}     & \multicolumn{4}{c|}{Pr($> \mid z \mid $)}                                              \\ \hline
                & Human  & Script & Linguistic & Base   & Human & Script & Linguistic & Base  & Human               & Script              & Linguistic          & Base                \\ \hline
(Intercept)     & -3.4   & -3.418 & -3.245     & -3.061 & 0.244 & 0.279  & 0.321      & 0.791 & \textless 2e-16 *** & \textless 2e-16 *** & \textless 2e-16 *** & 0.00011 ***         \\ \hline
recency         & 1.322  & 1.322  & 1.324      & 1.322  & 0.095 & 0.095  & 0.096      & 0.097 & \textless 2e-16 *** & \textless 2e-16 *** & \textless 2e-16 *** & \textless 2e-16 *** \\ \hline
frequency      & 0.097  & 0.103  & 0.112      & 0.114  & 0.098 & 0.097  & 0.098      & 0.102 & 0.317               & 0.289               & 0.251               & 0.262               \\ \hline
pastObj         & 0.407  & 0.396  & 0.423      & 0.395  & 0.293 & 0.294  & 0.295      & 0.3   & 0.165               & 0.178               & 0.151               & 0.189               \\ \hline
pastSubj        & -0.967 & -0.973 & -0.909     & -0.926 & 0.559 & 0.564  & 0.562      & 0.565 & 0.0838 .            & 0.0846 .            & 0.106               & 0.101               \\ \hline
pastExpPronoun  & 1.603  & 1.619  & 1.616      & 1.602  & 0.21  & 0.207  & 0.208      & 0.245 & 2.19e-14 ***        & 5.48e-15 ***        & 7.59e-15 ***        & 6.11e-11 ***        \\ \hline
depTypeSubj     & 2.939  & 2.942  & 2.656      & 2.417  & 0.299 & 0.347  & 0.429      & 1.113 & \textless 2e-16 *** & \textless 2e-16 *** & 5.68e-10 ***        & 0.02994 *           \\ \hline
depTypeObj      & 1.199  & 1.227  & 0.977      & 0.705  & 0.248 & 0.306  & 0.389      & 1.109 & 1.35e-06 ***        & 6.05e-05 ***        & 0.0119 *            & 0.525               \\ \hline
surprisal       & -0.04  & -0.006 & 0.002      & -0.131 & 0.099 & 0.097  & 0.117      & 0.387 & 0.684               & 0.951               & 0.988               & 0.735               \\ \hline
residualEntropy & -0.009 & 0.023  & -0.141     & -0.128 & 0.088 & 0.128  & 0.168      & 0.258 & 0.916               & 0.859               & 0.401               & 0.619               \\ \hline
\end{tabular}
}
\caption{Coefficients obtained from regression analysis for different models. Two NP types considered: full NP and Pronoun/ProperNoun, with base class full NP. Significance:  `***' $<0.001$, `**' $<0.01$, `*' $<0.05$, and `.' $<0.1$.}
\label{table:secondReg}
\end{table*}

\subsection{A model of Referring Expression Choice}

Our goal is to determine whether referent predictability (quantified in terms of surprisal) is
correlated with the type of referring expression used in the text. Here we
focus on the distinction between pronouns and full noun phrases. Our data also contains a small
percentage (ca.~1\%) of proper names (like ``John''). Due to this
small class size and earlier findings that proper nouns behave much
like pronouns \cite{tily2009refer}, we combined pronouns and proper
names into a single class of short encodings.

\abovedisplayskip=0pt
\belowdisplayskip=1pt
For the referring expression type prediction task, we estimate the surprisal of the referent from each of our computational models from Section \ref{sec:drmodel} as well as the human cloze task.  The surprisal of an upcoming discourse referent $d^{(t)}$ based on the previous context $h^{(t)}$ is thereby estimated as: 
\begin{equation}
\nonumber
\tiny
S(d^{(t)}) = - \log\ p(d^{(t)} \mid h^{(t)})%
\label{eq:surprisalDR}
\end{equation}
In order to determine whether referent predictability has an effect on referring expression type \textit{over and above} other factors that are known to affect the choice of referring expression, we train a logistic regression model with referring expression type as a response variable and discourse referent predictability 
as well as a large set of other linguistic factors (based on Tily and Piantadosi, 2009) as explanatory variables. The model is defined as follows: \\
\begin{equation}
\tiny
\nonumber
p(n^{(t)} = n|d^{(t)},h^{(t)}) =
\frac{\exp(\mathbf{v}^{T} \mathbf{g}(n, d^{t}, h^{(t)}))} {  \sum_{n'} { \exp(\mathbf{v}^{T} \mathbf{g}(n', d^{t}, h^{(t)})) }     },%
\end{equation}
\noindent where $d^{(t)}$ and $h^{(t)}$ are defined as before, $\mathbf{g}$ is
the feature function, and $\mathbf{v}$ is the vector of model
parameters. The summation in the denominator is over NP types (full NP
vs.~pronoun/proper noun).  



%

\subsection{RE Model Experiments}
We ran four different logistic regression models. These models all
contained exactly the same set of linguistic predictors but differed in
the estimates used for referent type surprisal and residual
entropy. One logistic regression model used surprisal estimates based on the human referent cloze task, while the
three other models used estimates based on the three computational models (Base, Linguistic and Script). 
For our experiment, we are interested in the choice of referring expression type for those occurrences of references, where a ``real choice" is possible. We therefore exclude for our analysis reported below all first mentions as well as all first and second person pronouns (because there is no optionality in how to refer to first or second person). This subset contains 1345 data points.
 
\subsection{Results}
The results of all four logistic regression models are shown in Table
\ref{table:secondReg}.  We first take a look at the results for the linguistic
features. While there is a bit of variability in terms of the exact
coefficient estimates between the models (this is simply due to small correlations between these predictors and the predictors for surprisal), the
effect of all of these features is largely consistent across models.
For instance, the positive coefficients for the recency feature means that when a previous mention happened very recently, the referring expression is more likely to be a pronoun (and not a full NP).

The coefficients for the surprisal estimates of the different models are, however, not significantly different from zero. Model comparison shows that they do not improve model fit. We also used the estimated models to predict referring expression type on new data and again found that surprisal estimates from the models did not improve prediction accuracy. This effect even holds for our human cloze data. Hence, it cannot be interpreted as a problem with the models---even human predictability estimates are, for this dataset, not predictive of referring expression type.

We also calculated regression models for the full dataset including first and second person pronouns as well as first mentions (3346 data points). The results for the full dataset are fully consistent with the findings shown in Table \ref{table:secondReg}: there was no significant effect of surprisal on referring expression type. 

This result contrasts with the findings by \newcite{tily2009refer}, who reported a significant effect of surprisal on RE type for their data. In order to replicate their settings as closely as possible, we also included residualEntropy as a predictor in our model (see last predictor in Table \ref{table:secondReg}); however, this did not change the results.



\section{Discussion and Future Work}
Our study on incrementally predicting discourse referents showed that script knowledge is a highly important factor in determining human discourse expectations. Crucially, the computational modelling approach allowed us to tease apart the different factors that affect human prediction as we cannot manipulate this in humans directly (by asking them to ``switch off" their common-sense knowledge).

By modelling common-sense knowledge in terms of event sequences and event participants, our model captures many more long-range dependencies than normal language models. The script knowledge is automatically induced by our model from crowdsourced scenario-specific text collections. 

In a second study, we set out to test the hypothesis that uniform information density affects referring expression type. This question is highly controversial in the literature: while Tily and Piantadosi (2009) find a significant effect of surprisal on referring expression type in a corpus study very similar to ours, other studies that use a more tightly controlled experimental approach have not found an effect of predictability on RE type \cite{stevenson1994thematic,fukumura2010choosing,rohde2014grammatical}. The present study, while replicating exactly the setting of T\&P in terms of features and analysis, did not find support for a UID effect on RE type. The difference in results between T\&P 2009 and our results could be due to the different corpora and text sorts that were used; specifically, we would expect that larger predictability effects might be observable at script boundaries, rather than within a script, as is the case in our stories.

A next step in moving our participant prediction model towards NLP applications would be to replicate our modelling results on automatic text-to-script mapping instead of gold-standard data as done here (in order to approximate human level of processing). Furthermore, we aim to move to more complex text types that include reference to several scripts. We plan to consider the recently published ROC Stories corpus \cite{mostafazadeh2016corpus}, a large crowdsourced collection of topically unrestricted short and simple narratives, as a basis for these next steps in our research.




\section*{Acknowledgments}

We thank the editors and the anonymous reviewers for their insightful suggestions. We would like to thank Florian Pusse for helping with the Amazon Mechanical Turk experiment. We would also like to thank Simon Ostermann and Tatjana Anikina for helping with the InScript corpus. This research was partially supported by the German Research Foundation (DFG) as part of SFB 1102 `Information Density and Linguistic Encoding', European Research Council (ERC) as part of ERC Starting Grant BroadSem (\#678254), the Dutch National Science Foundation as part of NWO VIDI 639.022.518, and the DFG once again as part of the MMCI Cluster of Excellence (EXC 284).



%


\bibliographystyle{acl2012}
\bibliography{refrences}

\end{document}